# Optimizing LPB Algorithms using Simulated Annealing

Dana Rasul Hamad[1]. Tarik A. Rashid[2*]

**Abstract.** Learner Performance-based Behavior using Simulated Annealing (LPBSA) is an improvement of the Learner Performance-based Behavior (LPB) algorithm. LPBSA, like LPB, has been proven to deal with single and complex problems. Simulated Annealing (SA) has been utilized as a powerful technique to optimize LPB. LPBSA has provided results that outperformed popular algorithms, like the Genetic Algorithm (GA), Particle Swarm Optimization (PSO), and even LPB. This study outlines the improved algorithm's working procedure by providing a main population and dividing it into Good and Bad populations, and then applying crossover and mutation operators. When some individuals were born in the crossover stage, they have to go through the mutation process. Between these two steps, we have applied SA using the Metropolis Acceptance Criterion (MAC) so as to accept only the best and useful individuals to be used in the next iteration. Finally, the outcomes demonstrate that the population is enhanced, leading to improved efficiency and validating the performance of LPBSA.

## 8.1 Introduction

In the realm of optimization, many algorithms display significant promise; however, they require greater efficiency and robustness to effectively solve complex problems and obtain better results. The LPB algorithm shows potential, while, with its exploitation and exploration, there is room to improve or enhance its overall performance. Metaheuristics are sophisticated approaches for addressing problems that have been created to address a variety of optimization issues. These algorithms have demonstrated efficacy in managing intricate and insolvable scenarios, providing an approximate resolution within a brief timeframe. In many disciplines, including engineering, medicine, corporate planning, and energy management, optimization is essential. In addition, metaheuristic algorithms are capable of making intelligent choices among a wide range of potential solutions [1]. Natural systems are the source of inspiration for some popular metaheuristic algorithms, such as the GA and PSO. The LPB algorithm was recently introduced by Rahman and Rashid (2021) [2] and is intended to handle both single and multi-objective situations. Rahman and Rashid's study illustrates how to apply LPB in a methodical manner to maximize and acquire optimal results. Researchers interested in creating, improving, or hybridizing the method can use the publication as a guide as well.

The primary solution focuses on the improvement of LPB algorithm to address its limitations. To do this we introduced LPBSA, which integrates the strengths of LPB with powerful techniques of SA to enhance its performance. By, striking a balance between exploring new possibilities and exploiting known ones, LPBSA provided as the best approach to the algorithm converge more efficiently to the optimal solution. The proposed algorithm is an improvement on LPB that expands on its capabilities. Utilizing SA as a potent method, LPBSA outperforms well-known algorithms like GAs and PSO in terms of LPB optimization. By applying crossover and mutation operations, splitting a main population into Good and Bad populations, and using SA with a MAC, the LPBSA technique selects only the best candidates for the subsequent iteration. By improving the population, this technique raises the algorithm's efficiency and improves its performance.

Optimization algorithms are crucial in terms of solving complex or high-dimensional problems effectively. Among those algorithms, LPB has obtained impressive results by drawing inspiration from the academic progress of students, especially students that are currently moving to university from high school. They are new students from college and have chosen a new field in which they may find it difficult to improve their performance, so they need to share information with other students in order to improve their behavior. To this end, the LPB algorithm tries to use its methods and techniques to improve the performance of students. LPB still has room to improve its performance, which can be achieved by integrating an additional metaheuristic technique to address its limitation. One of the main objectives of developing LPB algorithm is to enhance students' behavior with the goal of improving their overall experience across academic disciplines. Consequently, progressions in the LPB algorithm directly contribute to the enhancement of students' knowledge in various fields. Therefore, algorithms can be developed in order to iteratively refine their search processes, leading to more efficient solutions. In order to strengthen the LPB, simulating annealing was chosen suitably because it is a popular metaheuristic algorithm that was inspired by the metallurgical annealing process. In this process, materials are gradually cooled to obtain an optimal crystalline structure. Similarly, SA aids the algorithm escape local optima, making it extremely effective in improving the exploration phase of the optimization process. Also, SA increases the possibility of discovering the goal optimum because it helps in broadening the search space and probabilistically.

## 8.2 Literature Review

Optimization algorithms play an important role for the complex problem solutions through a wide range of fields. Such as, their application in education has garnered pivotal attention, as these algorithms provide potential solutions for improving learning outcomes, support decision-making and personalizing learning experience. Among these optimization approaches, the GA stands out due to its capability to effectively search large solution spaces, inspired by the principles of nature selection. GA has established to be essential in enhancing several tasks, from resources allocation to curriculum design, highlighting its versatility and important in educational contexts. This algorithm has also been applied to resource allocation and optimization scheduling in educational settings. For instance, GAs have been used to mimic adaptive systems for learning, which tailors educational content to the need of a specific learner [3]. PSO is another optimization algorithm that was inspired by both “fish schooling” and the “social behavior of flocking birds”, and it has been utilized for optimization tasks in numerus fields such as education. It is also absolutely useful in e-learning systems, especially for some significant facilities related to these systems, including improving the collaborative learning area and developing intelligent tutorials. One of PSO's aspects that makes it suitable for "real-time educational applications" is that this algorithm has a high ability to converge fast to optimal solutions [4]. Moreover, [5] presents differential evolution as another simple and effective optimization algorithm that has been utilized in solving many problems in the education field, particularly for some relevant educational facilities like managing resources, scheduling exams, and designing curriculums. Simplicity and robustness are the two essential aspects that make this algorithm a popular choice for educational tasks.

Some approaches like "data-driven analytics", "machine learning models", and "education data mining" are aiming to identify trends in student data for developing their involvement targets. [6]states that one of the key links that will improve a student's achievement and engagement is personalized learning by addressing student learning styles, preferences, and wants. In addition, by using real-time data, "adaptive learning systems" can be significantly utilized to alter the contents and learning environment to meet the learner student's needs. Moreover, optimization algorithms are employed by these systems to modify feedback, assessment, and instructional resources dynamically. Furthermore, the systems created based on these algorithms can lead to greater student fulfillment and outcomes for improved learning [7]. Next, more of a focus will be on existing gaps not addressed properly in previous studies.

In this section, we will highlight some significant gaps which will be focused in this study, especially regarding personalized learning and optimization algorithms. One of the crucial gap that has not been addressed properly is the challenge of admitting students into university programs for which they may lack foundational knowledge. This problem highlights the need for more effective algorithms that will assist the students to improve their skills (performance) by selecting their specific learning needs. By refining these algorithms, better supports will be available for the students in overcoming knowledge gaps and enhance their academic progress. One of the most important areas for tracking and exposing previous algorithms is “scalability”. Many optimization algorithms struggle during applications to large educational datasets. With existing algorithms, there is a need to provide actionable and interpretable insights for educators. This is another gap that should be under focus. Another gap is that, so far, there is a lack of focus on integrating optimization algorithms for real-world educational systems and practices. LPBSA, as a novel contribution, provides a scalable, interpretable, and integrative approach to address these gaps to optimize learner performance. LPBSA is inspired by the process of admitting students to the colleges or universities where they are encouraged to adapt and enhance their study skills or behaviors. This process aligns with the essential principles of personalized and adaptive learning, which focus on tailoring education to meet the student needs. By the incorporation of SA, which strengthens the LPB results, LPBSA has made the algorithm more reliable and, more importantly, it also designed to be adaptable and responsive for further use in the future, making it essential for optimizing processes and develop outcomes in various fields like healthcare and education.

## 8.3. LPB

The phrases required in learner acceptance are mirrored in the LPB algorithm, which draws inspiration from the university admissions process for recent high school graduates. These procedures include grouping students according to their cumulative rates and subsequently enhancing their behavior and performance after admission. These techniques also help to improve individuals’ conduct and performance level when they are admitted to a particular department. According to [8] [9], given the necessity for learners to adopt new study habits as they transition from junior high to college, in order to solve this, the algorithm first chooses a subset of the population. These individuals are then separated into smaller groups, from which the most qualified applications are selected in accordance with their qualifications. Then, through cooperative efforts that resemble cooperation and facilitate information exchange during study sessions, their conduct and performance are further improved (referred to as crossover). Furthermore, this approach introduces mutation, or unpredictable impacts, on their behavior. LPB incorporates mutation and crossover strategies similar to those used in GA. Researchers used the diagram shown in Fig 8. 1 as a flowchart to develop this algorithm.

It is important to represent how the LPB algorithm works in the next paragraph.

The initialization plays the first step in the process in which a group of people(learners) is separated into two subgroups based on their fitness. Subsequently, work targets at enhancing behavior through promoting working as a team and information transfer within the sub-groups. This approach along with the development of effective study habits, is aimed at improving individual performance and learning outcomes. Later, we have to select the best individuals from each sub-group based on their improved behaviors.

Then we start to exchange information between selected individuals, called “crossover”, to create new solutions. This is similar to exchanging information between parents to produce offspring. When crossover is complete, the “mutation” operation starts, which introduces random changes to some individuals to explore new potential solutions, and assess the fitness of the new individuals to determine their performance. Then, the process is repeated to select the next iteration. The process stops when a certain condition is met.

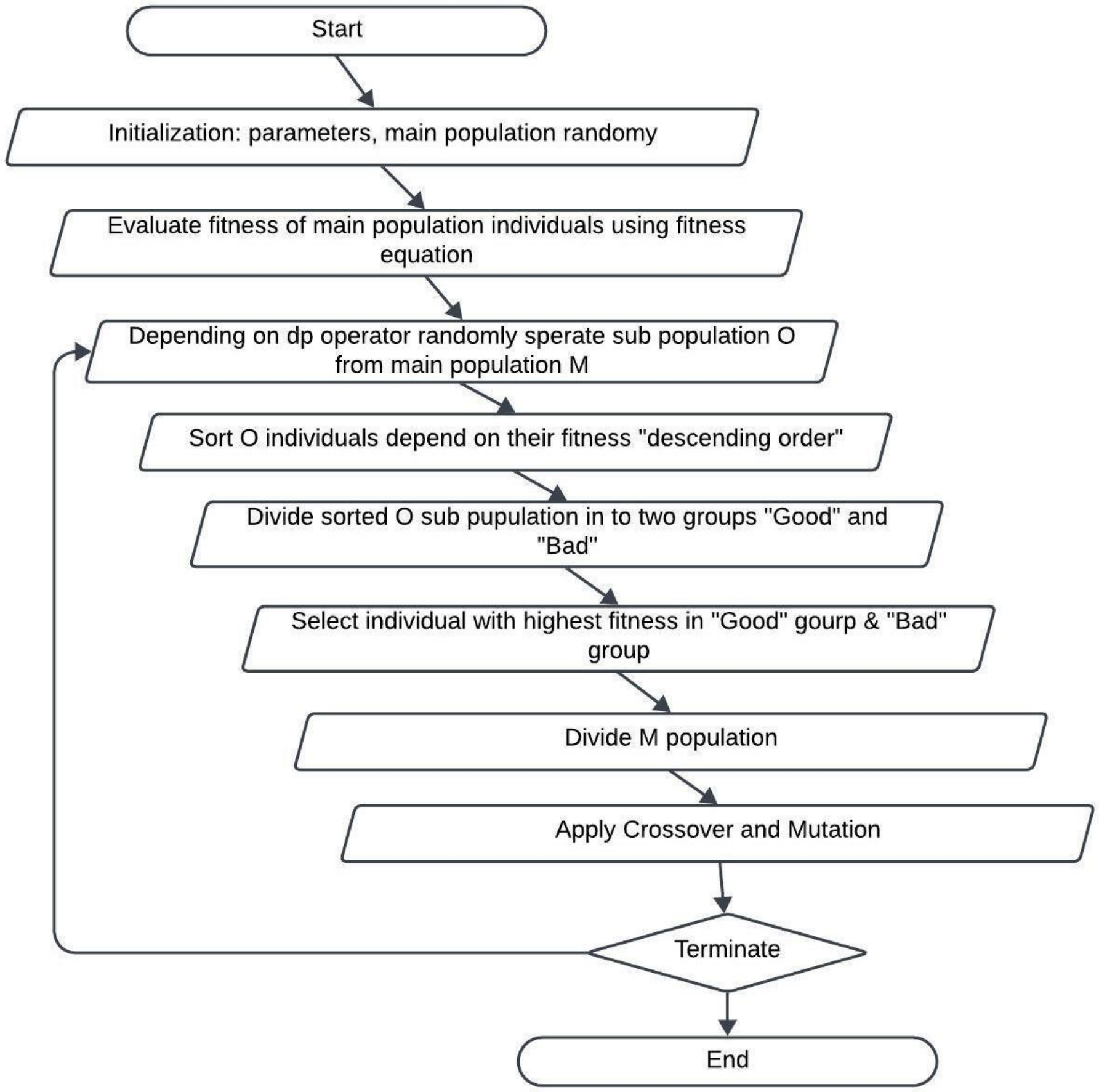

*Fig 8. 1: LPB Flowchart*

## 8.4. SA

The SA algorithm was introduced in 1983 by Kirkpatrick et al. to solve the "Traveling Salesman Problem (TSP)" [10]. SA is a probabilistic way to estimate a function's global optimum. In addition, it is a metaheuristic specifically designed to approximate global optimization for an optimization problem across a broad search space, and it is an optimization technique that maintains constant control limits while systematically modifying the variables, and upper and lower truncation limits in a stochastic way [11].

SA algorithms draw inspiration from metallurgy's annealing process, which involves rapidly heating a metal followed by gradual cooling [12]. At high temperatures, atoms within the metal exhibit fast movement, but as the temperature decreases, their kinetic energy diminishes. Ultimately, this process leads to a more ordered atomic arrangement, resulting in a material that is more ductile and easier to manipulate [12]. Numerous optimization issues, including TSP, protein folding, graph partitioning, and job-shop scheduling, have been effectively solved with SA [13] [14]. The capacity of SA to break free from local minima and converge to a global minimum is its primary benefit [11]. Furthermore, SA does not require prior knowledge of the search space and is quite simple to implement. Starting with an initial solution, the SA method iteratively improves the existing solution by randomly perturbing it and accepting the perturbation with a predetermined probability [15]. With more iterations, there is a progressive drop in the probability of accepting a worse answer from its original high level [15]. Fig 8. 2 offers a through flowchart that illustrates each phase of SA.

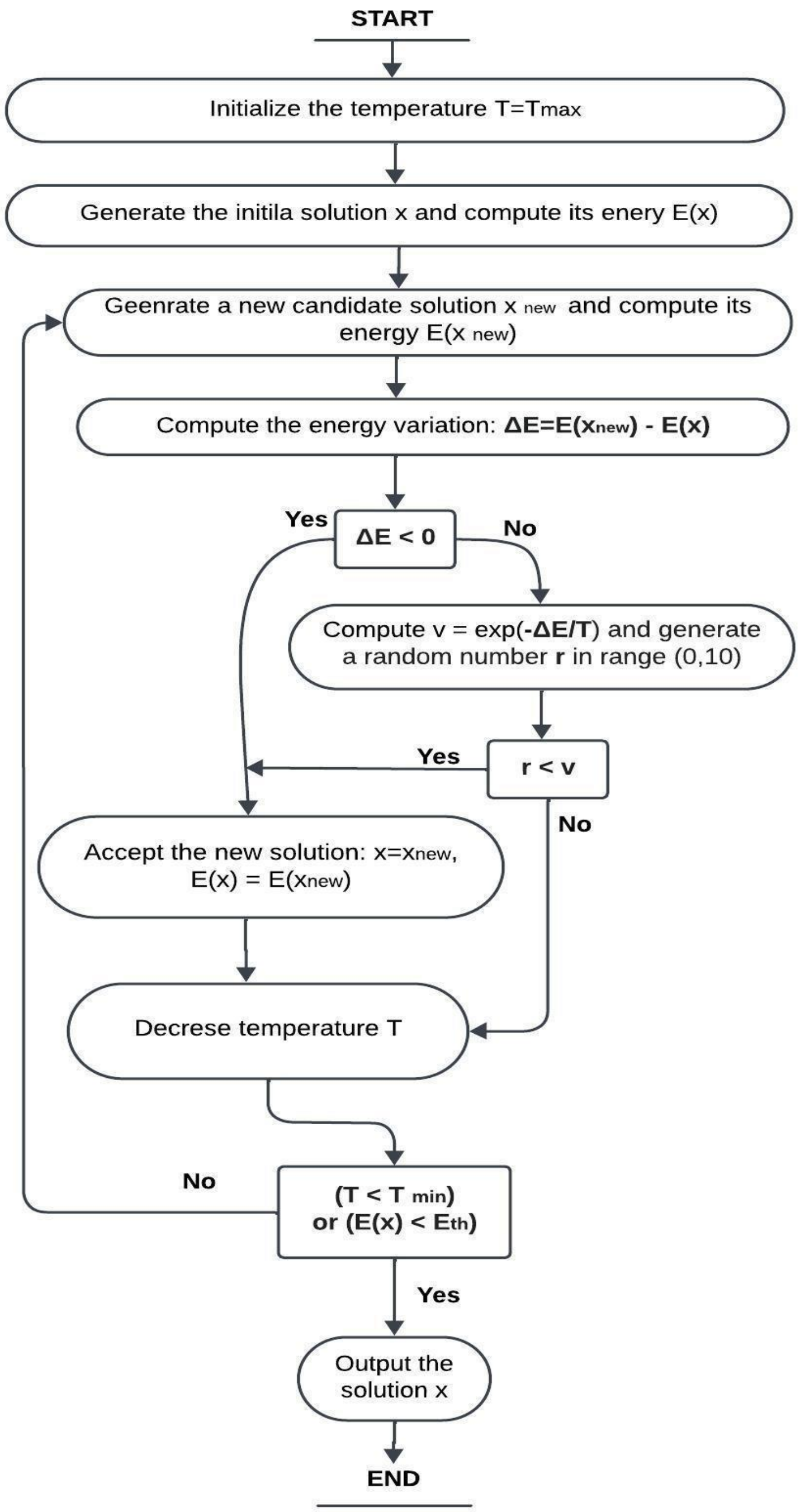

*Fig 8. 2: Simulated Annealing Algorithm Flowchart*

As mentioned, SA is an optimization method inspired by the annealing process in metallurgy. It involves heating a material and then cooling it slowly to remove defects, analogous to exploring and refining solutions in optimization.

### 8.4.1 Key Concepts Behind SA

Some key concepts have a significant role to play in developing the techniques of the SA algorithm. “Energy states and cost functions” is one of those concepts. In optimization, the energy concept in a system is analogous to the cost function, that shows the value required to be minimized or maximized. Each acceptable solution to the problem optimization corresponds to a state of the system. And, based on cost function the quality of that solution is evaluated. Also, temperature and cooling schedule are another two key concepts where temperatures acts as a controlling parameter to determine the likelihood of accepting less optimal solution across the algorithms progress, allowing the exploration of a broader solution space. Means with a higher probability of a higher temperature allow the algorithm to accept worse solutions, promoting exploration of the solution space. Based on a predefined cooling schedule, the temperature is regularly reduced. As the temperature reductions, the possibility of accepting worse solution is reduced, guiding the algorithm toward convergence at an optimal solution.

The Metropolis Acceptance Criterion is another concept and an important key to SA’s effectiveness that allows the acceptance of worse solutions developed on the basis of a probability that decreases with temperature assisting the algorithm in escaping local minima. LPBSA combines the robust exploration capabilities of SA with the structured approach of LPB. So, by incorporating the Metropolis Acceptance Criterion , LPBSA enhances its capacity to avoid local minima and explore the solution space more effectively.

### 8.4.2 How the algorithm works

In the context of the algorithm’s work process, here is a step-by-step presentation of how the algorithm works.

Initialization

Start the algorithm with a high temperature or cooling rate. This is known as the initial solution. Then the cost of the initial solution will be evaluated.

Iterative Process

The following steps are repeated until the specified number of iterations is reached or the stopping criteria are met:

- ✓ Generate a new solution: The existing solution is disrupted to produce a new candidate solution and then the cost of the new solution is evaluated.
- ✓ Evaluate the new solution: The difference between the new solution and the current solution in cost ($\Delta E$) is calculated.
- ✓ Apply the Metropolis Acceptance Criterion: The (MAC) determines the likelihood of accepting a solution with a higher cost, enabling the algorithm to explore a broader solution space.
- ✓ Update the temperature: According to the cooling schedule, reduce the temperature.

Convergence

The proposed algorithm concentrates on refining the solution discovered during the temperature decrease and then the algorithm becomes less inclined to accept worse solutions. When the temperature meets the stopping condition or gets close to zero, the algorithm converges to an optimum solution.

## 8.5 LPBSA

LPBSA is a hybrid optimization technique that blends SA and LPB. However, it is known as an improvement approach that focuses more on the LPB algorithm’s optimization. By increasing convergence, resilience, and adaptability in the solution of challenging optimization problems, LPBSA seeks to improve on established LPB approaches. By comparing LPBSA with LPB and other well-known algorithms, such as PSO and GA, LPBSA performs better through assessments utilizing benchmark test functions. LPBSA plays a crucial role in improving the efficiency of intelligent systems by fine-tuning parameters, lowering the complexity of computation, and improving performance overall. This algorithm especially helpful in dynamic settings where the optimal solution could alter over time. Because of its adaptive qualities, LPBSA can modify its search strategies to change conditions or circumstances.

### 8.5.1 Practical Implementation using LPBSA to Optimize Learner Performance

LPBSA holds essential possible for optimizing learner performance in diverse educational settings. In this section, this algorithm's capabilities are integrated into real-world environments. Essentially, given the purpose of personalized learning pathways, this algorithm provides promising chances in real-world educational settings to improve learner performance over personalized learning pathways by analyzing individual performance data, wherever students are struggling, and tailoring learning activities to their precise needs. The algorithm can identify such areas, which will have a great effect in terms of improving learning experiences. There are significant potential for the performance of optimizing learners are available in various education settings of LPBSA, It is also has an ability to effectively applied in real world environments to provide personalized learning pathways. LPBSA identifies areas where students are facing challenges and tailors learning activities to address their specific requirements based on analyzing individual performance data. As well as, this targeted approach improves learning skills and helps in enhance overall academic outcomes. One of the most prominent implementation considerations is "data collection", which means it is very significant to collect comprehensive data on student performance, such as grades, attendance, or any other relevant metrics. Optimized resource sharing and adaptive learning systems, and identifying students that are currently at-risk, can be beneficial for LPBSA, which is crucial because it will adjust the difficulty of lessons based on real-time feedback from the algorithm, dynamically ensuring that each student is challenged appropriately. In terms of implementation considerations, LPBSA should be integrated with learning management systems (LMS) in order to adjust learning materials automatically.

The primary objectives of this altered algorithm are to improve convergence, adaptability and robustness. By integrating LPB with SA the approach improves global exploration, serving to avoid local optima. And, introduces temperature-controlled evolution that allows the acceptance of less optimal solutions when beneficial and combines SA's exploration abilities with adaptive learning to create an effective optimization strategy. Which is ensuring robustness throughout several problem domain. The following section explains how LPBSA works.

LPBSA starts by initializing a population of solutions and calculating the fitness of each individual. It then selects a subset of individuals, sorts them into Good and Bad groups based on fitness, and assigns each individual to a group. Next, it selects individuals for crossover, ensuring the ideal group is not empty, and performs crossover to create new individuals. Mutation is applied to introduce random changes. The algorithm uses the Metropolis Acceptance Criterion to accept or reject new individuals based on a random number and temperature rate. The population is updated with the accepted individuals and the process is repeated for a specified number of iterations or until a termination condition is met. The optimal solution is returned at the end. Fig 8. 3 shows the flowchart of LPBSA.

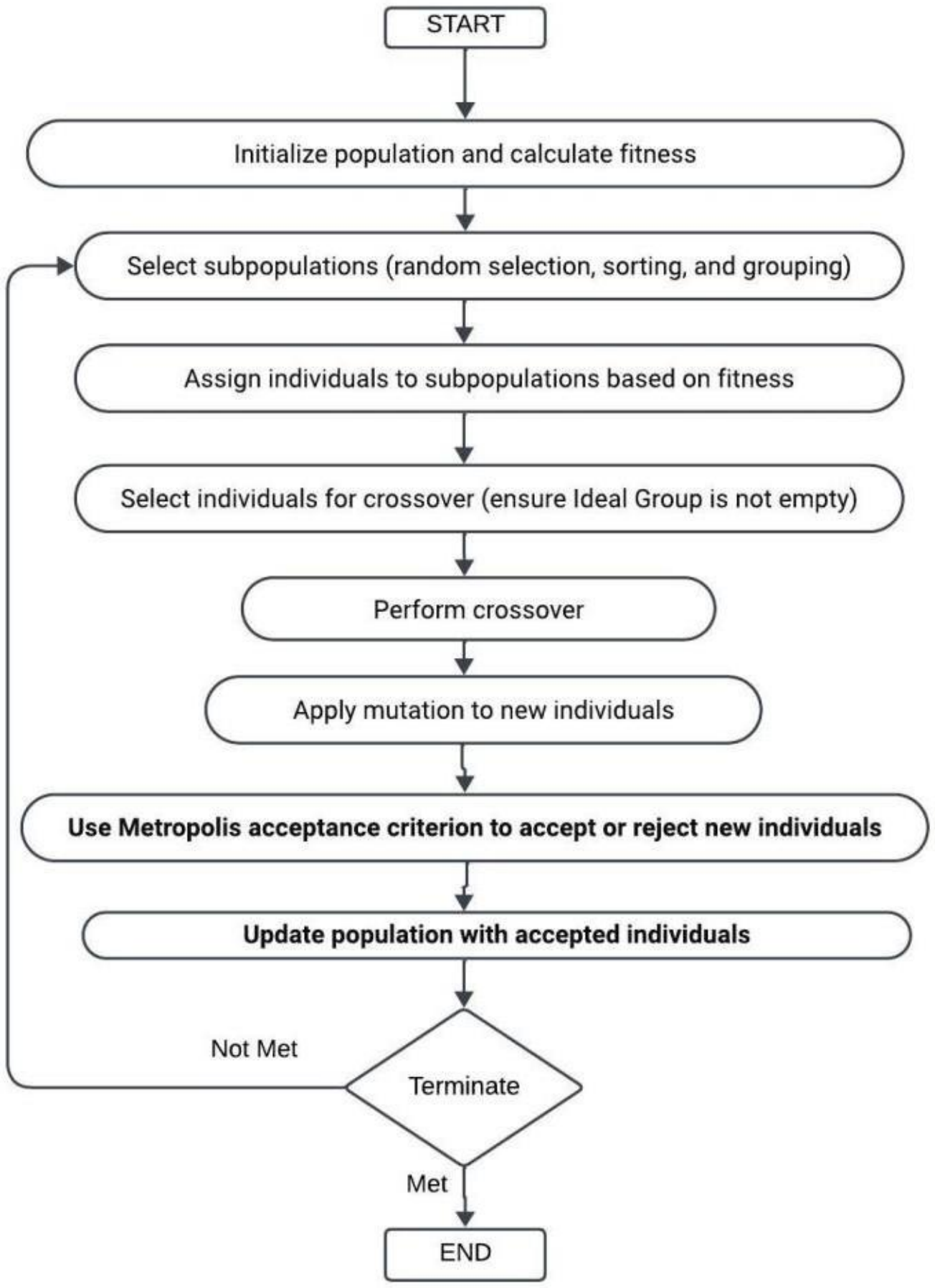

*Fig 8. 3: LPBSA Flowchart*

The Metropolis Acceptance Criterion is applied before proceeding to the next iteration in LPBSA during the mutation process. The criterion is defined in Equation 8. 1. By calculating the probability of accepting the worse solution, this equation helps in prevent premature convergence and maintain diversity within the population by calculating the probability of accepting less optimal solutions.

$$P(accept\ worse)\ exp\ (- \frac{Cost(new) - Cost(current)}{temperature})$$

*Equation 8. 1*

## 8.6 Key Steps of Crossover and Mutation in LPBSA

There are some reasons behind crossover and mutation operators being used with LPBSA, such as "exploration and exploitation". Crossover mainly enables exploration by sharing information from diverse regions in the search space. On the other hand, mutation facilitates exploitation by implementing random and small modifications, which fine-tune solutions. Hence, these two operators together balance exploitation and exploration, which is very significant in order to provide effective optimization. [16] states that, by properly balancing two modes like exploitation and exploration, increase the chances to achieve robust results. And, Diversity maintenance is important for preserving genetic diversity within the population, by leveraging both crossover and mutation. Which is allowing the algorithm to reduce the risk of getting trapped in local optima. So, this diversity is crucial for achieving global optima which is the ultimate goal of the optimization process.

Regarding “adaptability”, LPBSA can adapt to various optimization landscapes by combining crossover and mutation. The algorithms capability to fine-tune solutions within the fitness environment is ensured through the mutation process.

### 8.6.1 Crossover Operator

In LPBSA the crossover operator is inspired by the process of integrating the behaviors of individuals to deliver a new individual with possibly superior characteristics. In LPBSA, the crossover operator is utilized to combine information from two solutions to create a new candidate solution, aiding to explore the search space more effectively. Two parents are selected for crossover:

i. Parents are introduced as a decimal number that is converted to binary.
ii. The left half of the binary number for the first parent is merged with the right half of the binary number for the second parent, and the right half of the binary number for the first parent is integrated with the left half of the binary number for the second parent.
iii. Two new offspring are produced.

### 8.6.2 Mutation Operator

In LPBSA, the mutation operator provides a random change to the individual solution, which means more improvement in selected individuals. This promotes variety within the population and aids the avoidance of premature convergence. The mutation process is performed simply as presented below.

i. New offspring from crossover are selected for mutation.
ii. A binary number is utilized for modification.
iii. One point of the binary number is randomly changed, like one 0 point changed to 1.
iv. New and improved offspring are delivered.

## 8.7 Case Study

Consider the following function: $f(x) = x_1^2 + x_2^2$, for integer $x$, $0 \leq x_1 \leq 9000$ and $0 \leq x_2 \leq 9000$.

**Step 1**: Create a random population (Bi) of between 0 and 9000 individuals and then provide an average for the **fitness of $X_1^2$and $X_2^2$**, as shown in Table 8. 1.

*Table 8. 1: Main population (Bi)*

| **Bi** | $X_1$ | $X_2$ | $X_1^2$ | $X_2^2$ | **Fitness of $X_1^2$and $X_2^2$** |
|---|---|---|---|---|---|
| B1 | 4320 | 3120 | 18662400 | 9734400 | 28396800 |
| B2 | 1233 | 4523 | 1520289 | 20457529 | 21977818 |
| B3 | 5100 | 3209 | 26010000 | 10297681 | 36307681 |
| B4 | 4355 | 5210 | 18966025 | 27144100 | 46110125 |
| B5 | 2331 | 4266 | 5433561 | 18198756 | 23632317 |
| B6 | 2040 | 2755 | 4161600 | 7590025 | 11751625 |
| B7 | 5043 | 1977 | 25431849 | 3908529 | 29340378 |
| B8 | 3460 | 4781 | 11971600 | 22857961 | 34829561 |
| B9 | 1920 | 5510 | 3686400 | 30360100 | 34046500 |
| B10 | 4222 | 3741 | 17825284 | 13995081 | 31820365 |
| B11 | 5401 | 1740 | 29170801 | 3027600 | 32198401 |
| B12 | 3351 | 2850 | 11229201 | 8122500 | 19351701 |
| B13 | 5201 | 4989 | 27050401 | 24890121 | 51940522 |
| B14 | 2188 | 3477 | 4787344 | 12089529 | 16876873 |
| B15 | 3409 | 1877 | 11621281 | 3523129 | 15144410 |
| B16 | 4560 | 2776 | 20793600 | 7706176 | 28499776 |

Average: 28889053

**Step 2**: Create subpopulation (s) randomly from the main population by selecting eight individuals and put them in descending order. Divide those eight individuals into two groups (Good) and (Bad). Select individuals from highest to lowest from those groups, which is shown in Table 8. 2.

*Table 8. 2: K Population*

| **Bi** | **Ki** | **Fitness** | **Groups** |
|---|---|---|---|
| B3 | K1 | **36307681** | Good |
| B8 | K2 | 34829561 | |
| B11 | K3 | 32198401 | |
| B1 | K4 | 28396800 | |
| B5 | K5 | **23632317** | Bad |
| B12 | K6 | 19351701 | |
| B15 | K7 | 15144410 | |
| B6 | K8 | 11751625 | |

Highest fitness for the Good group = **36307681**

Highest fitness for the Bad group = **23632317**

**Step 3**: Compare the main population (*Bi*) individuals to the highest fitness of the Good and Bad group in Ki, which is presented in Table 8. 3.

- If the fitness values of Bi <= highest fitness of the Bad group, move Bi to the Bad group;
- else if the fitness value of Bi <= highest fitness of the Good group, move Bi to the Good group;

- else if the fitness value of Bi > highest fitness of the Good group, move Bi to the Ideal group.

*Table 8. 3: Comparison of Bi individuals with Ki highest Good and Bad fitness*

| Bi | Fitness | Fit (Bi)<= Fit (Ki5) | Fit (Bi)<= Fit(Ki1) | Ideal group |
|---|---|---|---|---|
| B1 | 28396800 | | Good | |
| B2 | 21977818 | Bad | | |
| B3 | 36307681 | | Good | |
| B4 | 46110125 | | | Ideal |
| B5 | 23632317 | Bad | | |
| B6 | 11751625 | Bad | | |
| B7 | 29340378 | | Good | |
| B8 | 34829561 | | Good | |
| B9 | 34046500 | | Good | |
| B10 | 31820365 | | Good | |
| B11 | 32198401 | | Good | |
| B12 | 19351701 | Bad | | |
| B13 | 51940522 | | | Ideal |
| B14 | 16876873 | Bad | | |
| B15 | 15144410 | Bad | | |
| B16 | 28499776 | | Good | |

**Step 4**: In this step we have to select 4 individuals from the main population, ensure that the Ideal group is not empty; if it is, choose individuals from the Good group. If the Good group is also empty, select individuals from the Bad group. Table 8. 4 displays the selected individuals from Table 8. 3, these four individuals are used for the next phase (Crossover stage).

Ensure that the number of selected individuals matches the specified requirement $N = 4$, as stated in the initial step.

*Table 8. 4: Selected Individuals*

| Proper individuals are selected to crossover from Table 8.3 | | |
|---|---|---|
| B13 | 51,940,522 | Ideal |
| B4 | 46,110,125 | Ideal |
| B3 | 36,307,681 | Good |
| B8 | 34,829,561 | Good |

**Step 5**: Table 8. 5 presents the applied crossover between the individuals selected from Table 8.4 with Ki individuals from Table 8.2.

*Table 8. 5: Crossover operation*

| **IdCh** | **NInd** | $X_1$ | $X_2$ | **Binary ($X_1$)** | **Binary($X_2$)** |
|---|---|---|---|---|---|
| E1 | B13 | 5201 | 4989 | **101000**1010001 | **100110**1111101 |
| E2 | K1 | 5100 | 3209 | 100111**1101100** | 110010**001001** |
| C1 | New | 5228 | 2441 | 1010001101100 | 100110001001 |
| C2 | New | 5073 | 6525 | 1001111010001 | 1100101111101 |
| E3 | B4 | 4355 | 5210 | **100010**0000011 | **101000**1011010 |
| E4 | K2 | 3460 | 4781 | 110110**000100** | 100101**0101101** |
| C3 | New | 2180 | 5165 | 100010000100 | 1010000101101 |
| C4 | New | 6915 | 4826 | 1101100000011 | 1001011011010 |
| E5 | B3 | 5100 | 3209 | **100111**1101100 | **110010**001001 |
| E6 | K3 | 3351 | 2850 | 110100**010111** | 101100**100010** |
| C5 | New | 2519 | 3234 | 100111010111 | 110010100010 |
| C6 | New | 6764 | 2825 | 1101001101100 | 101100001001 |
| E7 | B8 | 3460 | 4781 | **110110**000100 | **100101**0101101 |
| E8 | K4 | 4320 | 3120 | 100001**1100000** | 110000**110000** |
| C7 | New | 7008 | 2416 | 1101101100000 | 100101110000 |
| C8 | New | 1028 | 6189 | 10000000100 | 1100000101101 |

**Step 6**: For the new individuals (children), apply mutation and convert one 0 to 1 for each individual randomly in order to maximize the function. This process is shown in Table 8. 6.

*Table 8. 6: Apply Mutation*

| **IdCh** | $X_1$ | $X_2$ | **Binary ($X_1$)** | **Binary($X_2$)** |
|---|---|---|---|---|
| C1 | 5228 | 2441 | 1010001101100 | 100110001001 |
| C2 | 5073 | 6525 | 1001111010001 | 1100101111101 |
| C3 | 2180 | 5165 | 100010000100 | 1010000101101 |
| C4 | 6915 | 4826 | 1101100000011 | 1001011011010 |
| C5 | 2519 | 3234 | 100111010111 | 110010100010 |
| C6 | 6764 | 2825 | 1101001101100 | 101100001001 |
| C7 | 7008 | 2416 | 1101101100000 | 100101110000 |
| C8 | 1028 | 6189 | 10000000100 | 1100000101101 |

New individuals are shown in Table 8. 7.

*Table 8. 7: New individuals after Mutation*

| **IdCh** | $X_1$ | $X_2$ | **Binary ($X_1$)** | **Binary($X_2$)** |
|---|---|---|---|---|
| C1 | 7276 | 3465 | 1110001101100 | 110110001001 |
| C2 | 7121 | 7549 | 1101111010001 | 1110101111101 |
| C3 | 3204 | 7213 | 110010000100 | 1110000101101 |
| C4 | 7939 | 6874 | 1111100000011 | 1101011011010 |
| C5 | 3543 | 3746 | 110111010111 | 111010100010 |
| C6 | 7788 | 3849 | 1111001101100 | 111100001001 |
| C7 | 7024 | 3440 | 1101101110000 | 110101110000 |
| C8 | 1540 | 7213 | 11000000100 | 1110000101101 |

**Step 7**: Before passing to the next iteration, apply the (MAC) formula shown in Equation 8. 1. In this chapter, a random number = 0.6 and temperature rate = 100 are provided.

For each cost, compare the acceptance probability with the random number to determine acceptance or rejection:

For C1: Acceptance probability = exp(-(7276 - 5228)/100) ≈ 0.6703. Since 0.6 < 0.6703, C1 is **accepted**.

For C2: Acceptance probability = exp(-(7121 - 5073)/100) ≈ 0.6775. Since 0.6 < 0.6775, C2 is **accepted**.

For C3: Acceptance probability = exp(-(3204 - 2180)/100) ≈ 0.7408. Since 0.6 < 0.7408, C3 is **accepted**.

For C4: Acceptance probability = exp(-(7939 - 6915)/100) ≈ 0.6525. Since 0.6 < 0.6525, C4 is **accepted**.

For C5: Acceptance probability = exp(-(3543 - 2519)/100) ≈ 0.7288. Since 0.6 < 0.7288, C5 is **accepted**.

For C6: Acceptance probability = exp(-(7788 - 6764)/100) ≈ 0.6742. Since 0.6 < 0.6742, C6 is **accepted**.

For C7: Acceptance probability = exp(-(7024 - 7008)/100) ≈ 0.5391. Since 0.6 > 0.5391, C7 is **rejected**.

For C8: Acceptance probability = exp(-(1540 - 1028)/100) ≈ 0.6428. Since 0.6 < 0.6428, C8 is **accepted**.

Therefore, based on the comparison with the random number (0.6), C7 is rejected, whereas all other costs are accepted. Table 8. 8 shows the individuals that are accepted and used for the next iteration.

*Table 8. 8: Accepted Individuals*

| **IdCh** | $X_1$ | $X_2$ | **Binary (**$X_1$**)** | **Binary(**$X_2$**)** |
|---|---|---|---|---|
| C1 | 7276 | 3465 | 1110001101100 | 110110001001 |
| C2 | 7121 | 7549 | 1101111010001 | 1110101111101 |
| C3 | 3204 | 7213 | 110010000100 | 1110000101101 |
| C4 | 7939 | 6874 | 1111100000011 | 1101011011010 |
| C5 | 3543 | 3746 | 110111010111 | 111010100010 |
| C6 | 7788 | 3849 | 1111001101100 | 111100001001 |
| C8 | 1540 | 7213 | 11000000100 | 1110000101101 |

**Step 8**: Calculate the new accepted individuals that are displayed in Table 8. 9.

*Table 8. 9: Calculate accepted individuals*

| **IdCh** | $X_1$ | $X_2$ | $X_1^2$ | $X_2^2$ | **Fitness of** $X_1^2$ **and** $X_2^2$ |
|---|---|---|---|---|---|
| C1 | 7276 | 3465 | 52940176 | 12006225 | 64946401 |
| C2 | 7121 | 7549 | 50708641 | 56987401 | 107696042 |
| C3 | 3204 | 7213 | 10265616 | 52027369 | 62292985 |
| C4 | 7939 | 6874 | 63027721 | 47251876 | 110279597 |
| C5 | 3543 | 3746 | 12552849 | 14032516 | 26585365 |
| C6 | 7788 | 3849 | 60652944 | 14814801 | 75467745 |
| C8 | 1540 | 7213 | 2371600 | 52027369 | 54398969 |

**Step 9**: Calculate the average for accepted individuals in Table 8.9, the individuals in Table 8.4 and (Good group) individuals in Table 8.2. The results are shown in Table 8. 10.

*Table 8. 10: Provide an average*

| **IdCh** | $X_1$ | $X_2$ | $X_1^2$ | $X_2^2$ | **Fitness of** $X_1^2$ **and** $X_2^2$ |
|---|---|---|---|---|---|
| B13 | 5201 | 4989 | 27050401 | 24890121 | 51940522 |
| K1 | 5100 | 3209 | 26010000 | 10297681 | 36307681 |
| B4 | 4355 | 5210 | 18966025 | 27144100 | 46110125 |
| K2 | 3460 | 4781 | 11971600 | 22857961 | 34829561 |
| B3 | 5100 | 3209 | 26010000 | 10297681 | 36307681 |
| K3 | 3351 | 2850 | 11229201 | 8122500 | 19351701 |
| B8 | 3460 | 4781 | 11971600 | 22857961 | 34829561 |
| K4 | 4320 | 3120 | 18662400 | 9734400 | 28396800 |
| C1 | 7276 | 3465 | 52940176 | 12006225 | 64946401 |
| C2 | 7121 | 7549 | 50708641 | 56987401 | 107696042 |
| C3 | 3204 | 7213 | 10265616 | 52027369 | 62292985 |
| C4 | 7939 | 6874 | 63027721 | 47251876 | 110279597 |
| C5 | 3543 | 3746 | 12552849 | 14032516 | 26585365 |
| C6 | 7788 | 3849 | 60652944 | 14814801 | 75467745 |
| C8 | 1540 | 7213 | 2371600 | 52027369 | 54398969 |

Average: 52649382

Steps 2–9 are repeated until the desired number of iterations or the termination condition is satisfied, at which point the optimal solution is provided. Let's follow Steps 2–9 for the next iteration.

**Step 2**: From the main population *B*, create a new subpopulation *K* which are illustrated in Table 8. 11.

*Table 8. 11: Second K Sub-population*

| **Bi** | **K** | **Fitness** | **Groups** |
|---|---|---|---|
| B4 | K1 | **46110125** | Good |
| B9 | K2 | 34046500 | |
| B10 | K3 | 31820365 | |
| B1 | K4 | 28396800 | |
| B2 | K5 | **21977818** | Bad |
| B12 | K6 | 19351701 | |
| B14 | K7 | 16876873 | |
| B6 | K8 | 11751625 | |

Highest fitness for the Good group = **46110125**

Highest fitness for the Bad group = **21977818**

**Step 3**: Compare the main population (*Bi*) individuals to the highest fitness of the Good and Bad group in *K*, presented in Table 8. 12.

- If the fitness values of *Bi* <= highest fitness of the Bad group, move *Bi* to the Bad group;
- else if the fitness value of *Bi* <= highest fitness of the Good group, move *Bi* to the Good group;
- else if the fitness value of *Bi* > highest fitness of the Good group, move *Bi* to the Ideal group.

*Table 8. 12: Compare Bi individuals with Ki highest Good and Bad fitness*

| **Bi** | **Fitness** | **Fit (Bi)<= Fit (Ki5)** | **Fit (Bi)<= Fit(Ki1)** | **Ideal group** |
|---|---|---|---|---|
| B1 | 28396800 | | Good | |
| B2 | 21977818 | | Good | |
| B3 | 36307681 | | Good | |
| B4 | 46110125 | | Good | |
| B5 | 23632317 | | Good | |
| B6 | 11751625 | Bad | | |
| B7 | 29340378 | | Good | |
| B8 | 34829561 | | Good | |
| B9 | 34046500 | | Good | |
| B10 | 31820365 | | Good | |
| B11 | 32198401 | | Good | |
| B12 | 19351701 | Bad | | |
| B13 | 51940522 | | | Ideal |
| B14 | 16876873 | Bad | | |
| B15 | 15144410 | Bad | | |
| B16 | 28499776 | | Good | |

**Step 4**: : In this step we have to select 4 individuals from the main population, ensure that the Ideal group is not empty; if it is, choose individuals from the Good group. If the Good group is also empty, select individuals from the Bad group. The selected individuals are used in the next phase (Crossover Operator). The results are shown in Table 8. 13.

Ensure that the number of the selected individuals matches the specified requirement $N = 4$, as stated in the initial step.

*Table 8. 13: Selected individuals*

| **Proper individuals are selected to crossover from Table 8.13** | | |
|---|---|---|
| B13 | 51940522 | Ideal |
| B4 | 46110125 | Good |
| B3 | 36307681 | Good |
| B8 | 34829561 | Good |

**Step 5**: Apply crossover between the individuals selected from Table 8. 13 and Good group individuals from Table 8. 11. The results of this operation are shown in Table 8. 14.

*Table 8. 14: Crossover Operator*

| **IdCh** | **NInd** | $X_1$ | $X_2$ | **Binary ($X_1$)** | **Binary($X_2$)** |
|---|---|---|---|---|---|
| E1 | B13 | 5201 | 4989 | **101000**1010001 | **100110**1111101 |
| E2 | K1 | 4355 | 5210 | 100010**0000011** | 1010001**011010** |
| C1 | New | 5123 | 4954 | 1010000000011 | 1001101011010 |
| C2 | New | 4433 | 5245 | 1000101010001 | 1010001111101 |
| E3 | B4 | 4355 | 5210 | **100010**0000011 | **101000**1011010 |
| E4 | K2 | 1920 | 5510 | 111100**00000** | 101011**0000110** |
| C3 | New | 1088 | 5126 | 10001000000 | 1010000000110 |
| C4 | New | 7683 | 5594 | 1111000000011 | 1010111011010 |
| E5 | B3 | 5100 | 3209 | **101011**0000110 | **110010**001001 |
| E6 | K3 | 4222 | 3741 | 100000**1111110** | 111010**011101** |
| C5 | New | 5630 | 3229 | 1010111111110 | 110010011101 |
| C6 | New | 4102 | 3721 | 1000000000110 | 111010001001 |
| E7 | B8 | 3460 | 4781 | **110110**000100 | **100101**0101101 |
| E8 | K4 | 4320 | 3120 | 1000011**100000** | 110000**110000** |
| C7 | New | 3488 | 2416 | 110110100000 | 100101110000 |
| C8 | New | 4292 | 6189 | 1000011000100 | 1100000101101 |

**Step 6**: For the new individuals (children), apply mutation and convert one 0 to 1 for each individual randomly in order to maximize the function. This process is shown in Table 8. 15.

*Table 8. 15: Apply Mutation*

| **IdCh** | $X_1$ | $X_2$ | **Binary ($X_1$)** | **Binary($X_2$)** |
|---|---|---|---|---|
| C1 | 5123 | 4954 | 1010000000011 | 1001101011010 |
| C2 | 4433 | 5245 | 1000101010001 | 1010001111101 |
| C3 | 1088 | 5126 | 10001000000 | 1010000000110 |
| C4 | 7683 | 5594 | 1111000000011 | 1010111011010 |
| C5 | 5630 | 3229 | 1010111111110 | 110010011101 |
| C6 | 4102 | 3721 | 1000000000110 | 111010001001 |
| C7 | 3488 | 2416 | 110110100000 | 100101110000 |
| C8 | 4292 | 6189 | 1000011000100 | 1100000101101 |

New individuals are shown in Table 8. 16.

*Table 8. 16: New individuals after Mutation*

| **IdCh** | $X_1$ | $X_2$ | **Binary ($X_1$)** | **Binary($X_2$)** |
|---|---|---|---|---|
| C1 | 7171 | 7002 | 1110000000011 | 1101101011010 |
| C2 | 6481 | 7293 | 1100101010001 | 1110001111101 |
| C3 | 1600 | 7174 | 11001000000 | 1110000000110 |
| C4 | 7939 | 7642 | 1111100000011 | 1110111011010 |
| C5 | 7678 | 3741 | 1110111111110 | 111010011101 |
| C6 | 6150 | 3977 | 1100000000110 | 111110001001 |
| C7 | 4000 | 3440 | 111110100000 | 110101110000 |
| C8 | 6340 | 7213 | 1100011000100 | 1110000101101 |

**Step 7**: Before passing to the next iteration, apply the MAC formula shown in Equation 8. 1. In this chapter, we provide random number = 0.6 and temperature rate = 100. The results are the same as below:

For C1: Acceptance probability = exp(-(7171 - 5123)/100) ≈ 0.6703. Since 0.6 < 0.6703, C1 is **accepted**.

For C2: Acceptance probability = exp(-(6481 - 4433)/100) ≈ 0.6703. Since 0.6 < 0.6703, C2 is **accepted**.

For C3: Acceptance probability = exp(-(1600 - 1088)/100) ≈ 0.6065. Since 0.6 > 0.6065, C3 is **rejected**.

For C4: Acceptance probability = exp(-(7939 - 7683)/100) ≈ 0.6311. Since 0.6 > 0.6311, C4 is **rejected**.

For C5: Acceptance probability = exp(-(7678 - 5630)/100) ≈ 0.7061. Since 0.6 < 0.7061, C5 is **accepted**.

For C6: Acceptance probability = exp(-(6150 - 4102)/100) ≈ 0.6703. Since 0.6 < 0.6703, C6 is **accepted**.

For C7: Acceptance probability = exp(-(4000 - 3488)/100) ≈ 0.6055. Since 0.6 > 0.6055, C7 is **rejected**.

For C8: Acceptance probability = exp(-(6340 - 4292)/100) ≈ 0.6703. Since 0.6 < 0.6703, C8 is **accepted**.

Based on the comparison with the random number (0.6), C3, C4, and C7 are rejected, whereas all other costs are accepted, Table 8. 17 shows the accepted individuals.

*Table 8. 17: Accepted individuals*

| **IdCh** | $X_1$ | $X_2$ | **Binary (**$X_1$**)** | **Binary(**$X_2$**)** |
|---|---|---|---|---|
| C1 | 7171 | 7002 | 1110000000011 | 1101101011010 |
| C2 | 6481 | 7293 | 1100101010001 | 1110001111101 |
| C5 | 7678 | 3741 | 1110111111110 | 111010011101 |
| C6 | 6150 | 3977 | 1100000000110 | 111110001001 |
| C8 | 6340 | 7213 | 1100011000100 | 1110000101101 |

**Step 8**: Calculate the new accepted individuals after applying simulated annealing the results are shown in Table 8. 18.

*Table 8. 18: Calculate accepted individuals*

| **IdCh** | $X_1$ | $X_2$ | $X_1^2$ | $X_2^2$ | **Fitness of** $X_1^2$ **and** $X_2^2$ |
|---|---|---|---|---|---|
| C1 | 7171 | 7002 | 51423241 | 49028004 | 100451245 |
| C2 | 6481 | 7293 | 42003361 | 53187849 | 95191210 |
| C5 | 7678 | 3741 | 58951684 | 13995081 | 72946765 |
| C6 | 6150 | 3977 | 37822500 | 15816529 | 53639029 |
| C8 | 6340 | 7213 | 40195600 | 52027369 | 92222969 |

**Step 9**: Provide the sum, average, and max for the accepted individuals in Table 8. 18 as well as the selected individuals from Table 8. 13. These results are displayed in Table 8. 19.

*Table 8. 19: Provide sum, average, and max*

| **IdCh** | $X_1$ | $X_2$ | $X_1^2$ | $X_2^2$ | **Fitness of** $X_1^2$ **and** $X_2^2$ |
|---|---|---|---|---|---|
| E1 | 5201 | 4989 | 27050401 | 24890121 | 51940522 |
| E2 | 4355 | 5210 | 18966025 | 27144100 | 46110125 |
| E3 | 4355 | 5210 | 18966025 | 27144100 | 46110125 |
| E4 | 1920 | 5510 | 3686400 | 30360100 | 34046500 |
| E5 | 5100 | 3209 | 26010000 | 10297681 | 36307681 |
| E6 | 4222 | 3741 | 17825284 | 13995081 | 31820365 |
| E7 | 3460 | 4781 | 11971600 | 22857961 | 34829561 |
| E8 | 4320 | 3120 | 18662400 | 9734400 | 28396800 |
| C1 | 7171 | 7002 | 51423241 | 49028004 | 100451245 |
| C2 | 6481 | 7293 | 42003361 | 53187849 | 95191210 |
| C5 | 7678 | 3741 | 58951684 | 13995081 | 72946765 |
| C6 | 6150 | 3977 | 37822500 | 15816529 | 53639029 |
| C8 | 6340 | 7213 | 40195600 | 52027369 | 92222969 |

**Average:** 55693299

Therefore, it is evident from the outcomes that the population has been enhanced, leading to increased efficiency among individuals. Table 8. 20 presents a comparison of the average for each iteration.

*Table 8. 20: Comparision of Averages*

| **Iteration *N*** | **Average** |
|---|---|
| Iteration 0 | 28889053 |
| Iteration 1 | 52649382 |
| Iteration 2 | 55693299 |

Fig 8. 4 shows the average results for each iteration and how they increased from iteration 0 to 2.

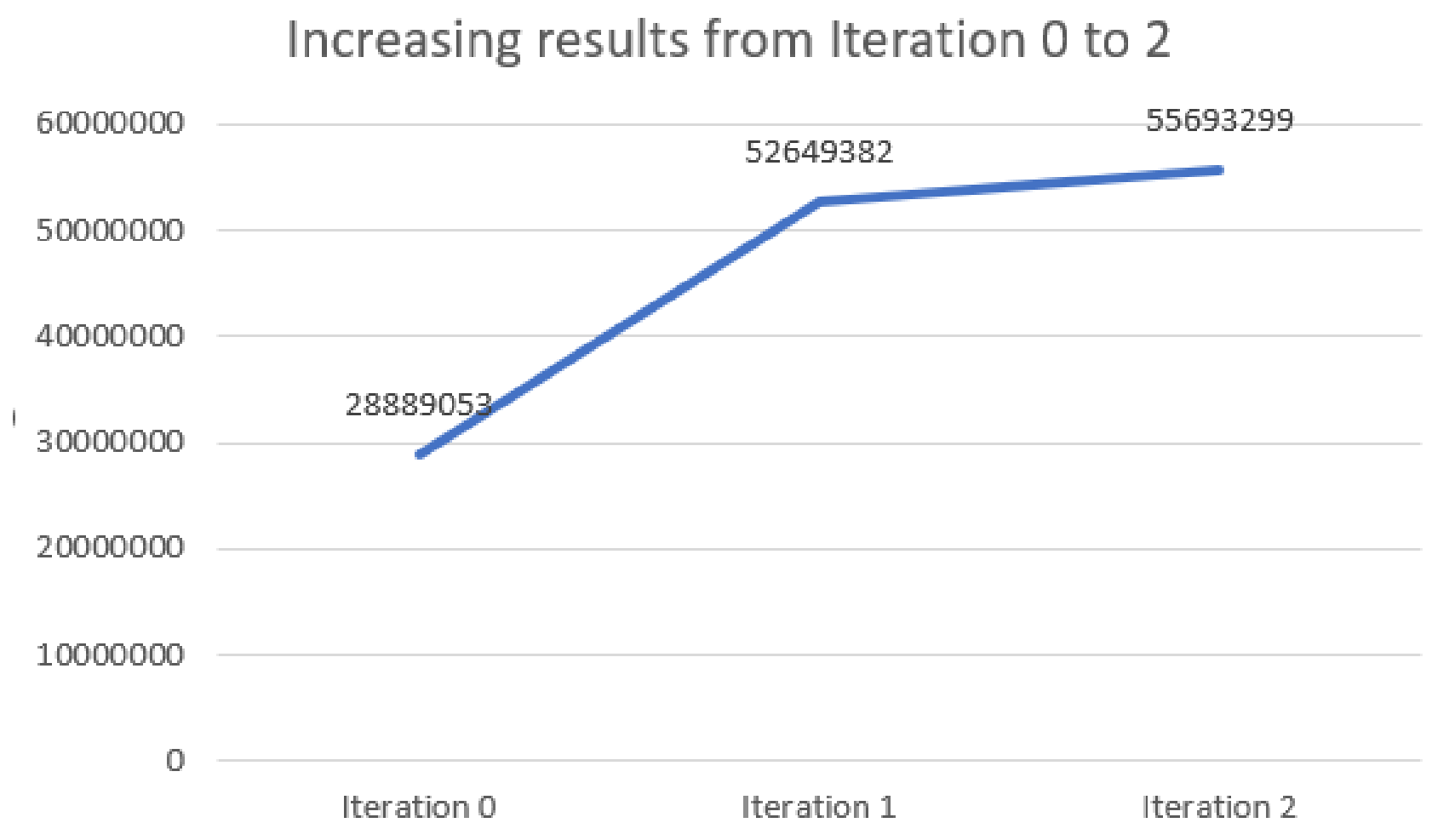

*Fig 8. 4: Average results of three iterations*

## 8.8 Comparative Analysis of LPBSA with Other Algorithms

As mentioned, LPBSA claims to outperform other algorithms. Therefore, it is crucial to present a detailed comparison analysis utilizing particular benchmarks. In this section, more focus is on the comparative analysis, the rationale behind the selected benchmarks, and the detailed outcomes that were obtained.

### 8.8.1 Metrics for Comparison

The metrics discussed in this section have played a major role in determining the performance of LPBSA. The average value is one of those metrics that can be utilized to represent the cost function average value that was obtained over multiple runs, signifying the algorithm’s capability to deliver promised solutions consistently. To reflect the algorithm’s reliability and stability, as well as measure the cost function value variability, standard deviation is utilized as another metric. Additionally, the next metric is the best value that represent the lowest cost function value, to assess the algorithm's capability to find the optimal solution. The worst value is used as another metric to obtain the highest cost function value to display the worst case of the algorithm’s performance.

### 8.8.2 Benchmark Test Functions

To evaluate the performance of LPBSA, 19 benchmark test functions were utilized. These functions are commonly used to measure the efficiency of optimization algorithms due to their potential characteristics, such as dimensionality, separability, and multimodality. For example, test function one, which is a sphere function, is a simple unimodal function with a single global minimum, and test function two is a Rastrigin function used as a multimodal function with numerous local minima, inspiring the algorithm’s exploration capability.

These test functions provide a comprehensive evaluation of the algorithm’s performance. Different aspects of those algorithms are tested via those functions, such as their ability to converge to the global minima, avoid the local minima, and maintain stability. Table 8. 21 shows a summary of evaluating LPBSA with other algorithms. The table entries represent the average and standard deviation of the cost function value over 30 runs for each entry.

*Table 8. 21: Result of 19 benchmark test functions to compare LPBSA with other algorithms*

| TF | LPBSA | | LPB | | DA | | PSO | | GA | |
|---|---|---|---|---|---|---|---|---|---|---|
| | **AVA** | **STD** | **AVA** | **STD** | **AVA** | **STD** | **AVA** | **STD** | **AVA** | **STD** |
| TF1 | 3.86896E-04 | 7.25127E-04 | 0.001877 | 0.0020936 | **2.85E-18** | 7.16E-18 | 4.20E-18 | 4.31E-18 | 748.5972 | 324.926 |
| TF2 | 3.9134E-03 | 2.67553E-03 | 0.005238 | 0.0036525 | 1.49E-05 | 3.76E-05 | 0.003154 | 0.009811 | 5.971358 | 1.53310 |
| TF3 | 15.5732633 | 9.35452E+00 | 36.47488 | 29.224155 | 1.29E-06 | 2.10E-06 | 0.001891 | 0.003311 | 1949.003 | 994.273 |
| TF4 | 0.15603627 | 3.49740E-02 | 0.393866 | 0.135818 | 0.000988 | 0.002776 | 0.001748 | 0.002515 | 21.16304 | 2.60540 |
| TF5 | **4.76762333** | **2.77755315** | 16.76919 | 22.19251 | 7.600558 | 6.786473 | 63.45331 | 80.12726 | 133307.1 | 85007.6 |
| TF6 | 0.001353802 | 1.83590E-03 | 0.002031 | 0.0027832 | 4.17E-16 | 1.32E-15 | **4.36E-17** | **1.38E-16** | 563.8889 | 229.699 |
| TF7 | 0.002900520 | **0.001495889** | 0.004975 | 0.002965 | 0.010293 | 0.010293 | 0.005973 | 0.003583 | 0.166872 | 0.07257 |
| TF8 | -3723.968593 | 191.566968 | -3747.65 | **189.0206** | -2857.58 | 383.6466 | -7.10E+11 | 1.2E+12 | **-3407.25** | 164.478 |
| TF9 | **0.00067658** | **0.0007894** | 0.001567 | 0.001842 | 16.01883 | 9.479113 | 10.44724 | 7.879807 | 25.51886 | 6.66936 |
| TF10 | 0.01168585 | 6.82155E-03 | 0.017933 | 0.013532 | 0.23103 | 0.487053 | 0.280137 | 0.601817 | 9.498785 | 1.27139 |
| TF11 | 0.062534300 | 2.58492E-02 | 0.066355 | 0.030973 | 0.193354 | 0.073495 | 0.083463 | 0.035067 | 7.719959 | 3.62607 |
| TF12 | 3.06720E-05 | 5.69906E-05 | 2.79E-05 | 3.84E-05 | 0.031101 | 0.098349 | **8.57E-11** | **2.71E-10** | 1858.502 | 5820.21 |
| TF13 | 2.63645E-04 | 7.24801E-04 | 0.000309 | 0.000512 | 0.002197 | 0.004633 | 0.002197 | 0.004633 | 68047.23 | 87736.7 |
| TF14 | **0.998000000** | **4.51681E-16** | 0.998004 | 1.26E-11 | 103.742 | 91.24364 | 150 | 135.4006 | 130.0991 | 21.3203 |
| TF15 | **0.001032395** | **6.76267E-04** | 0.002358 | 0.003757 | 193.0171 | 80.6332 | 188.1951 | 157.2834 | 116.0554 | 19.1935 |
| TF16 | **-1.031600000** | **6.77522E-16** | -1.03163 | 2.46E-06 | 458.2962 | 165.3724 | 263.0948 | 187.1352 | 383.9184 | 36.6053 |
| TF17 | 2.705400000 | **1.35504E-15** | **0.397888** | 3.16E-06 | 596.6629 | 171.0631 | 466.5429 | 180.9493 | 503.0485 | 35.7940 |
| TF18 | **3.000000000** | **0.00000E+00** | 3.000142 | 0.000283 | 229.9515 | 184.6095 | 136.1759 | 160.0187 | 118.438 | 51.0018 |
| TF19 | **-3.862800** | **3.16177E-15** | -3.86278 | 9.61E-07 | 679.588 | 199.4014 | 741.6341 | 206.7296 | 544.1018 | 13.3016 |

## 8.9 Results and Discussion

The results in Table 8.21 show that LPBSA consistently achieved better performance than the other algorithms, like PSO, across the majority of benchmark functions, GA, and the original LPB. In addition, LPBSA achieved a lower standard deviation and superior average solution, and it also showed its ability to deliver better solutions more consistently. Table 8.21 presents a whole map for all the tests. The best tests are highlighted and the table shows the best outcomes of LPBSA. Some test function results are illustrated in the following sentences. For example, in TF1, LPBSA recorded an average of 3.87E-04 and standard deviation of 7.25E-4, outperforming LPB, which achieved an average of 1.88E-03 and standard deviation of 2.09E-03. PSO and GA provided lower results. PSO achieved 4.20E-18, but with a high standard deviation, and GA recorded an average of 748.60.

In TF5, LPBSA obtained an average of 4.7676233 and standard deviation of 2.7775532. By recording these results, LPBSA outperformed all the other proposed algorithms in this test function. Moreover, LPBSA recorded the highest average and standard deviation compared to other algorithms in TF9,TF14,TF15,TF16,TF18, and TF19. Even in TF17, LPBSA provided the highest standard deviation of 1.35504E-15, and achieved the highest average by recording 2.705400000 compared to other algorithms, except LPB, which recorded an average of 1.35504E-15. These results show the precision, consistency, robustness, and superior performance of LPBSA, showcasing the algorithm's enhanced capability to find the optimal solution with great stability.

## 8.10 Conclusion

LPBSA offers a promising approach to optimizing solutions in various domains. By integrating SA and LPB, LPBSA effectively balances exploration and exploitation, allowing for the discovery of high-quality solutions. LPBSA's ability to adapt to changing environments and its efficient search capabilities make it a valuable tool for solving complex optimization problems. In addition , LPBSA outperformed the LPB algorithm in terms of LPB optimization, indicating a notable improvement over the LPB algorithm. A total of 19 benchmark test functions were utilized and tested to compare the proposed algorithms. LPBSA beat well-known algorithms like PSO and GAs by utilizing SA as a potent optimization strategy. This study's systematic methodology, which includes mutation, crossover, population division, and SA techniques, successfully increased the population and boosted overall efficiency. Thus, these outcomes highlight LPBSA's potential as a useful tool for solving challenging issues in a variety of fields.